\documentclass[10pt,twocolumn,letterpaper]{article}

\usepackage[pagenumbers]{cvpr} 

\usepackage{graphicx}
\usepackage{amsmath}
\usepackage{amssymb}
\usepackage{booktabs}
\usepackage{verbatim}
\usepackage{lipsum}  
\usepackage{graphicx}
\usepackage{enumitem}
\usepackage{multirow}
\setlist{nosep}
\usepackage{makecell}
\usepackage{diagbox}
\usepackage{color}
\definecolor{cvprblue}{RGB}{0,123,255} 

\usepackage[pagebackref,breaklinks,colorlinks,citecolor=cvprblue]{hyperref}

\usepackage[capitalize]{cleveref}
\crefname{section}{Sec.}{Secs.}
\Crefname{section}{Section}{Sections}
\Crefname{table}{Table}{Tables}
\crefname{table}{Tab.}{Tabs.}

\title{CoBra: Complementary Branch Fusing Class and Semantic Knowledge for Robust Weakly Supervised Semantic Segmentation}

\author{
Woojung Han \quad Seil Kang \quad Kyobin Choo \quad Seong Jae Hwang\thanks{Corresponding author}\\ 
Yonsei University
\\{\tt\small \{dnwjddl, seil, chu, seongjae\}@yonsei.ac.kr}
}

\begin{document}

\newcommand{\ourmodel}{CoBra}
\newcommand{\lossfunction}{CLL}
\newcommand{\vv}{\mathbf{v}}

\maketitle

\begin{abstract}
    Leveraging semantically precise pseudo masks derived from image-level class knowledge for segmentation, namely image-level Weakly Supervised Semantic Segmentation (WSSS), remains challenging. While Class Activation Maps (CAMs) using CNNs have steadily been contributing to the success of WSSS, the resulting activation maps often narrowly focus on class-specific parts (e.g., only the face of a human). On the other hand, recent works based on vision transformers (ViT) have shown promising results based on their self-attention mechanism to capture the semantic parts but fail in capturing complete class-specific details (e.g., entire body parts of humans but also with a dog nearby). In this work, we propose \textbf{Co}mplementary \textbf{Bra}nch (\textbf{CoBra}), a novel dual branch framework consisting of two distinct architectures that provide valuable complementary knowledge of class (from CNN) and semantic (from ViT) to each branch. In particular, we learn Class-Aware Projection (CAP) for the CNN branch and Semantic-Aware Projection (SAP) for the ViT branch to explicitly fuse their complementary knowledge and facilitate a new type of extra patch-level supervision.
    Our model, through \textbf{CoBra}, fuses CNN and ViT's complementary outputs to create robust pseudo masks that integrate both class and semantic information effectively. Extensive experiments qualitatively and quantitatively investigate how CNN and ViT complement each other on the PASCAL VOC 2012 and MS COCO 2014 dataset, showing a state-of-the-art WSSS result. This includes not only the masks generated by our model but also the segmentation results derived from utilizing these masks as pseudo labels. 
\end{abstract}

\let\thefootnote\relax\footnote{Project 
page and Code:~\href{https://micv-yonsei.github.io/cobra2024}{Link}}
\vspace{-15pt}

\section{Introduction}
\vspace{-5pt}
\label{sec:intro}
\begin{figure}[t]
    \centering
    \includegraphics[width=1\columnwidth]{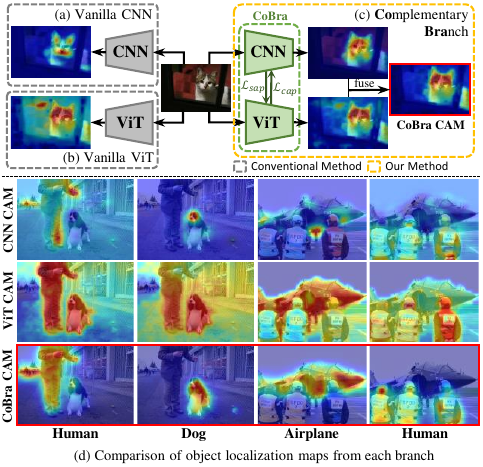}
    \vspace{-20pt}
    \caption{
    Illustration of the novel \textbf{Co}mplementary \textbf{Bra}nch framework that synergizes the class knowledge of CNN with the semantic understanding of ViT: (a) shows the standard CNN processing focused on class knowledge; (b) depicts the standard ViT utilizing semantic knowledge; (c) presents our integrated approach, where CNN and ViT branches exchange knowledge complementarily; and (d) compare  object localization maps from each branch. CNN, ViT, and Cobra branches for various subjects (\texttt{human, dog, airplane}), illustrating the distinctive areas of interest each model identifies. Our model successfully utilizes complementary characteristics to localize the exact object of the correct class and its semantic parts.}
    \label{fig:figure1}
\vspace{-25pt}
\end{figure}
%%%%%%%%%%%%%%%%%%%%%%%%%%%%%%%%%%%%%%%%%%%%
Among an array of contemporary computer vision tasks, semantic segmentation particularly continues to face difficulty acquiring accurate pixel-level segmentation labels which require a considerable amount of laborious human supervision. In response to such dependence on costly full supervision, weakly supervised semantic segmentation (WSSS) seeks to leverage other relatively less costly labels as \textit{weak} supervisions such as points \cite{points, gao2022weakly}, scribbles \cite{scribbles, zhang2020weakly}, and bounding boxes \cite{boundingbox1,boundingbox2,dong2021boosting}. Of particular interest is using \textit{image-level} labels (e.g., object class in image) which are already prevalent in existing vision datasets such as PASCAL VOC \cite{everingham2010pascal}. In this work, we specifically discuss WSSS using the existing image-level class label as weak supervision.
Unlike other types of weak supervision which naturally imply the object locations, the image-level class label only tells \textit{what} the objects are but does not tell \textit{where} they are. 
Thus, a line of prior works for (image-level) WSSS explored ways to utilize Class Activation Map (CAM) for generating object localization maps from CNN features \cite{cam, gradcam, gradcam++, advcam}. 
As shown in the first row of Fig.~\ref{fig:figure1}d, the objects are accurately localized by using the provided image-level class label (e.g., \texttt{human}, \texttt{dog} and \texttt{airplane}), primarily concentrating on the parts which contribute most to the classification task (e.g., face of \texttt{cat} in Fig.~\ref{fig:figure1}a). While this provides a strong class-specific localization cue, the resulting pixel-level pseudo label of the class often insufficiently covers the entire semantic region of the object (e.g., full body of \texttt{cat} in Fig.~\ref{fig:figure1}b). 
Interestingly, the recent surge of transformer-based models, namely vision transformer (ViT)~\cite{ViT}, demonstrated their high fidelity localization maps \cite{TS-CAM,mctformer,afa}.
Unlike CNN's CAM, ViT's localization map uses the self-attention mechanism to distinctly capture patch semantics, as seen in the middle row of Fig.~\ref{fig:figure1}d.
While this results in a sufficient coverage of the semantically sensible object regions (e.g., full body of \texttt{cat} in Fig.~\ref{fig:figure1}b) with accurate object boundaries, discriminating multiple classes, and separating foreground from background are particularly challenging due to the lack of strong class-specific cues \cite{TS-CAM} (e.g., \texttt{cat} and \texttt{window} in Fig.~\ref{fig:figure1}b).

In this work, we actively look into such \textit{complementary} characteristics of the aforementioned localization maps: class-specific localization map from CNN (e.g., Fig.~\ref{fig:figure1}d first row) and semantic-aware localization map from ViT (e.g., Fig.~\ref{fig:figure1}d middle row). To this end, we propose a novel \textbf{Complementary Branch (\ourmodel{})} consisting of a CNN branch and ViT branch which effectively reciprocate the complementary knowledge to each other. This dual branch framework enjoys the best of both worlds thus resulting in a localization map with a semantically precise boundary of correctly identified object (e.g., Fig.~\ref{fig:figure1}d last row). 

In practice, as illustrated in Fig.~\ref{fig:figure1}c, properly integrating the knowledge of these two completely different architectures (thus unshared weights) with distinct inductive biases requires special attention beyond the simple regularization of feature space. Related research also has been explored~\cite{howdovitwork}, where it analytically investigates the complementarity between Multi-head Self-Attention (MSA) and Convolutional Layers (Conv) and modifies the structural design of the model itself to integrate the strengths of both architectures. In contrast, we suggest that the mutual complementarity between the features of ViT and CNN can be facilitated using contrastive loss. To derive accurate representations of desired class and semantic features, we construct Class-Aware Projection (CAP) and Semantic-Aware Projection (SAP) from CNN and ViT respectively. Specifically, the CAP representation of each CNN CAM patch holds the class-specific information about the patch which is subsequently informed by the ViT attention map.
Conversely, the SAP representation of each ViT attention map holds the semantic relations between patches which are subsequently informed by the CNN CAM features. This well-constructed knowledge properly guides each branch to minimize its shortcomings (i.e., CNN lacking semantic sensitivity and ViT lacking class specificity) to generate class-and-semantic aware localization maps. To be more precise, the ViT-based semantic attention map tends to associate semantically relevant patches, while often disregarding the class (e.g., Fig.~\ref{fig:figure1}d shows that each objects are semantically well highlighted without knowing which class they belong to). 

\vspace{3px}
\noindent\textbf{Contributions.} We provide the following contributions:
\begin{itemize}[leftmargin=*]
    \item We propose a dual branch framework, namely \ourmodel{}, which aims to fuse the complementary nature of CNN and ViT localization maps.
    \item We capture the class and semantic knowledge as Class-Aware Projection (CAP) and Semantic-Aware Projection (SAP) respectively for effective complementary guidance to the CNN and ViT branches in \ourmodel{}, employing contrastive learning for enhanced guidance.
    
    \item We test our model's image-level WSSS performance on PASCAL VOC 2012 dataset and MS COCO 2014 and analyze the complementary relationship of class and semantic localization maps.
\end{itemize}
\vspace{-5pt}

\section{Related Works}
\vspace{-5pt}
\label{sec:related_works}
\subsection{CAM and CNN for WSSS}
\vspace{-7pt}

For leveraging the image-level class label as weak supervision, CAM \cite{cam} has been the most widely used approach for generating pseudo masks given image-level class labels. In particular, various CNN-based WSSS approaches have leveraged CAM for generating the initial seeds for pseudo labels \cite{seam, l2g, eps, ppc} and develop post-process techniques to improve the following mask \cite{sec, dsrg, irnet, psa, affinitynet}. Methods further focus on expanding the existing seed to sufficiently cover the entire object \cite{sec, dsrg}. Similarly, additional semantic information has been utilized to propagate the class activation throughout the relevant areas  \cite{irnet, psa, affinitynet}. Further, \cite{advcam, adversarialerasing} manipulated the input images in an adversarial manner to expand the seed by referring to obtained CAMs. To integrate additional weak supervision, the off-the-shelf saliency map was also considered as extra weak supervision \cite{l2g, eps}. As seen above, a line of work observed that the CNN-based CAM provides strong localization maps, but they primarily operate as \textit{seeds} which narrowly highlight only the most crucial object parts from the classification perspective.

\begin{figure*}[t!]
    \centering
    \includegraphics[width=1\textwidth]{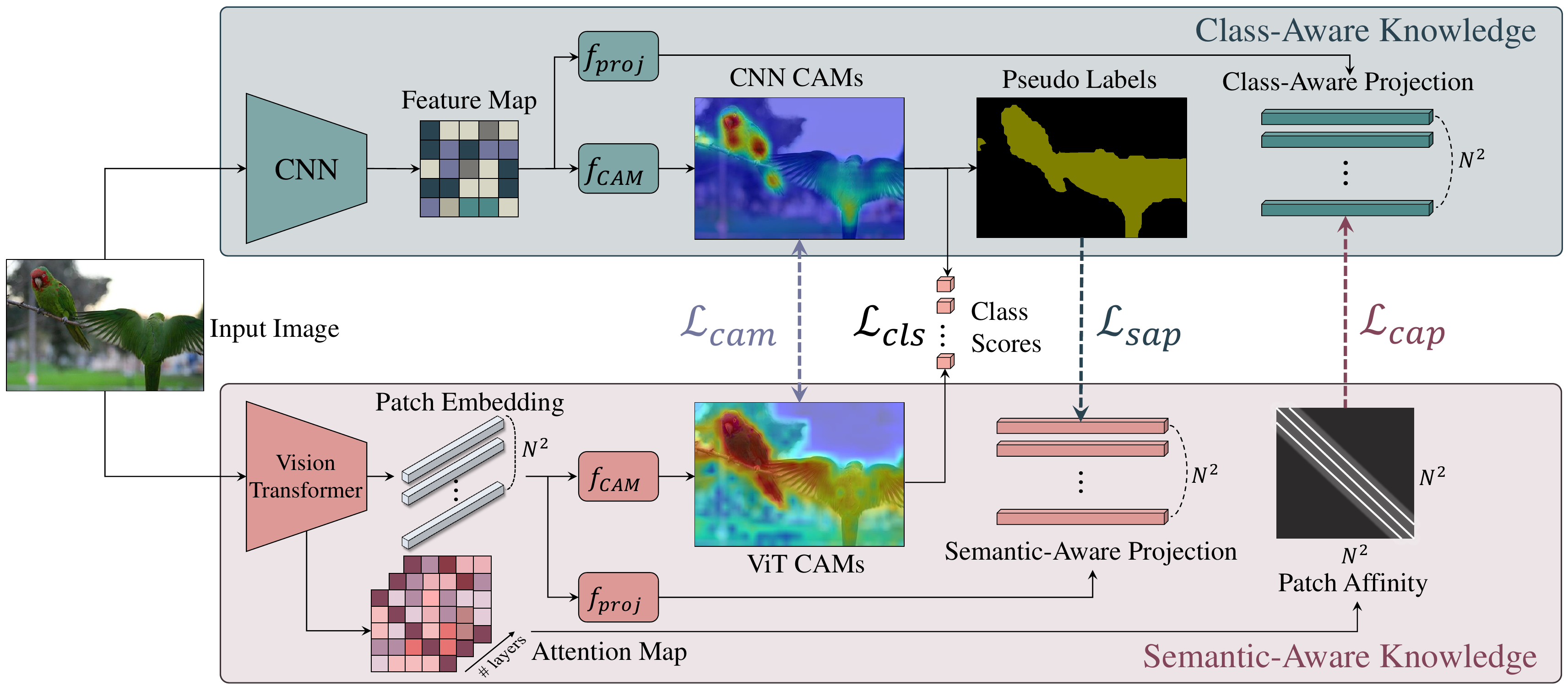}
    \vspace{-15pt}
    \caption{Overview illustration of our model, \textbf{Cross Complementary Branch (\ourmodel{})}. The dual branch framework consists of the Class-Aware Knowledge branch with CNN (top) and the Semantic-Aware Knowledge branch with ViT (bottom). The input image is passed down to both branches. \textbf{Class-Aware Knowledge (CAK) Branch}: The CNN outputs a feature map which generates (1) CNN CAMs via $f_{CAM}$, (2) Pseudo-Labels from CNN CAMs via $argmax$, and (3) Class-Aware Projection (CAP) via $f_{proj}$.
    \textbf{Semantic-Aware Knowledge (SAK) Branch}: The ViT outputs $N^2$ Patch Embeddings which generate (1) ViT CAMs via $f_{CAM}$ and (2) Semantic-Aware Projection (SAP) via $f_{proj}$. We also use the Attention Maps of all $L$-layers to generate (3) Patch Affinity of size $N^2 \times N^2$.
    \textbf{Complementary Branch Losses}: Once the necessary outputs are prepared, we employ various losses: (1) $\mathcal{L}_{cls}$: The typical classification loss based on the individual classification predictions of each CNN CAM and ViT CAM. (2) $\mathcal{L}_{cam}$: The L1 loss between the CNN CAM and ViT CAM. (3) $\mathcal{L}_{sap}$: The class-aware knowledge from the pseudo labels guides SAP to identify more accurate class-specific patches. (4) $\mathcal{L}_{cap}$: The semantic-aware knowledge from the patch affinity improves the semantic sensitivity of ViT CAM.
    }
    \label{fig:our_model}
\vspace{-10pt}
\end{figure*}

\vspace{-4pt}
\subsection{ViT in WSSS}
\vspace{-8pt}
Originating from Transformer \cite{transformer} in natural language processing, vision transformer (ViT) \cite{ViT} has been making notable breakthroughs in traditional computer vision tasks \cite{swintransformer, DETR}. Analogous to the tokens from words, ViT partitions an image into patches to construct their tokens, and its self-attention mechanism effectively captures the patch-wise relations within the image. This unique mechanism provides strong semantic cues of objects which is particularly useful for WSSS \cite{TS-CAM, mctformer, afa}. For instance, TS-CAM \cite{TS-CAM} extracts the attention map which highlights the regions of strong attention to the classifying object. Similarly, AFA \cite{afa} is an end-to-end framework using ViT as a backbone which refines the initial pseudo labels via semantic affinity from multi-head self-attention. These attention maps, however, cannot derive an attention map for \textit{each} class since the underlying ViT is based on a single, generic class token. MCTformer \cite{mctformer} proposed a multi-class token framework to generate a class-specific object localization map. These challenges still tend to include background and incorrect class.

\vspace{-5pt}
\subsection{CNN \& ViT in WSSS}
\vspace{-7pt}
Given the proven complementary relationship between CNN and ViT\footnote{For brevity, we begin referring vision transformer also as ViT.}, as demonstrated in prior research~\cite{howdovitwork}, it's not surprising that subsequent works have begun integrating both architectures. For instance, a straightforward hybrid approach either uses the embedding output of one architecture as an input to the other \cite{conformer} or simply uses the CNN to compute the CAM based on the token outputs of the ViT \cite{mctformer}.
On the other hand, there exist few prior works which explicitly attempt to make use of the localization maps of CNN and ViT. In particular, \cite{conformer, transcam} used multi-branch of CNN and transformer to retain the representation capability of local features and global representations to the maximum extent. Still, despite the impressive methodological developments, how such two models with vastly distinct inductive biases may bring complementary benefits toward image-level WSSS is unclear.

\vspace{-5pt}

\section{Methods} 
\vspace{-2pt}
\label{sec:methods}
\vspace{-8pt}
%%%%%%%%%%%%%%%%%%%%%% 3. METHODS  %%%%%%%%%%%%%%%%%%%%%%%
%%%%%%%%%%%%%%%%%%%%%%%%%%%%%%%%%%%%%%%%%%%%%%%%%%%%%%%%%%
In this section, we describe our model \ourmodel{} (shown in Fig.~\ref{fig:our_model}). In Sec.~\ref{sec:method1}, we provide an overview of our model motivating the dual branch setup consisting of the Class-Aware Knowledge (CAK) Branch (Fig.~\ref{fig:our_model} top) and the Semantic-Aware Knowledge (SAK) Branch (Fig.~\ref{fig:our_model} bottom). In Sec.~\ref{sec:method2}, for each branch, we describe its architecture and various outputs required for complementary knowledge infusion. In Sec.~\ref{sec:method3}, we formulate the complementary branch losses involving the Class-Aware Projection (CAP) and Semantic-Aware Projection (SAP). Finally, in Sec.~\ref{sec:method4}, a straightforward fusion method is employed to seamlessly integrate different knowledge elements, leading to the detailed description of the final object localization map essential for pseudo mask generation.
%%%%%%%%%%%%%%%%%%%%%%%%%%%%%%%%%%%%%%%%%%%%%%%%%%%%%%%%%%

%%%%%%%%%%%%%% 3.1. Overview & Motivation  %%%%%%%%%%%%%%%
% \vspace{-5pt}
\subsection{Overview and Motivation}
\vspace{-6pt}
\label{sec:method1}
We first motivate our Complementary Branch (CoBra) setup consisting of the Class-Aware Knowledge (CAK) Branch with CNN and the Semantic-Aware Knowledge (SAK) Branch with ViT.
%%%%%%%%%%%%%%%%%%%%%%%%%%%%%%%%%%%%%%%%%%%%%%%%%%%%%%%%%%

\vspace{4px}
\noindent\textbf{Class-Aware Knowledge Branch (CAK).} Shown as the top branch in Fig.~\ref{fig:our_model}, CNN is the basis of various components which possess strong class-related cues. For instance, the CNN CAMs accurately localize the object with very few incorrect localized regions (e.g., in Fig.~\ref{fig:figure1}a with low class false positives). In other words, the CNN-based CAK branch demonstrates \textit{high class precision}. Contrary to this benefit, the CNN CAMs tend to lack sufficient coverage of the remaining semantically relevant object parts (i.e., low semantic true positives). 
That is, CAK branch, on its own, has \textit{low semantic sensitivity} for generating pixel-level pseudo labels; thus requires additional semantic cues.
%%%%%%%%%%%%%%%%%%%%%%%%%%%%%%%%%%%%%%%%%%%%%%%%%%%%%%%%%%

\vspace{3px}
\noindent\textbf{Semantic-Aware Knowledge Branch (SAK).} Shown as the bottom branch in Fig.~\ref{fig:our_model}, ViT and its self-attention mechanism are the basis of semantically strong representations. For instance, the semantically persistent object regions are thoroughly captured in the ViT CAMs (e.g., in Fig.~\ref{fig:figure1}b with vivid boundaries of the localized objects). In other words, the ViT-based SAK branch demonstrates \textit{high semantic sensitivity}.
Converse to this valuable trait, we also observe a considerable size of falsely localized areas (i.e., high class false positives) such as the background and incorrect classes. In other words, SAK branch, on its own, has \textit{low class precision} for generating precise localization maps of specific objects; thus requires additional class cues.
%%%%%%%%%%%%%%%%%%%%%%%%%%%%%%%%%%%%%%%%%%%%%%%%%%%%%%%%%%
\begin{figure*}[t!]
    \centering
    \includegraphics[width=0.95\textwidth]{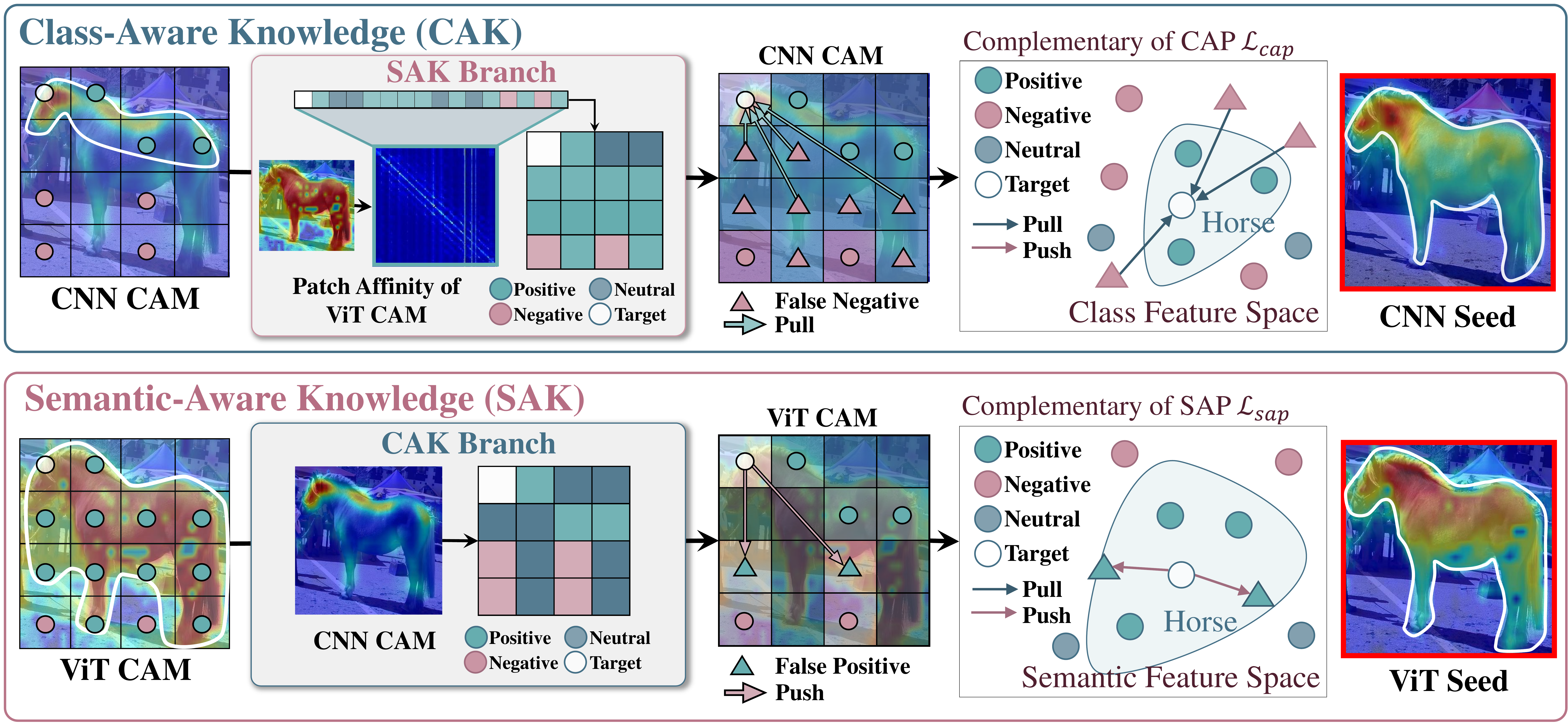}
    \caption{Illustration of refining CAP and SAP from SAK and CAK branch respectively.
    \textbf{(I) Class Aware Knoweldge(CAK):} The CAP values are embedded in the \textit{Class Feature Space}. (1) The CNN CAM shows that the false negative patches have been weakly localized as \texttt{horse}. (2) The patch affinity from SAK branch assigns the positive (green), negative (red), and neutral (teal) patches based on the target (white) patch. (3) The CAP loss (Eq.~\eqref{CAPloss}) pull those weakly localized patches (i.e., false class negatives) since they are assigned as semantically positive patches based on SAK branch. (4) The CAP is refined to improve the CNN CAM showing fewer false class negatives.
    \textbf{(II) Semantic Aware Knowledge(SAK):}
    The SAP values are embedded in the \textit{Semantic Feature Space}. (1) The ViT CAM shows that the negative patches have been incorrectly localized as \texttt{horse}. (2) The CNN CAM from CAK branch assigns the positive (green), negative (red), and neutral (teal) patches based on the target (white) patch. (3) The SAP loss (Eq.~\eqref{SAPloss}) pushes away those incorrectly localized patches (i.e., false class positives) since they are assigned as negative patches based on CAK branch. (4) The SAP is refined to improve the ViT CAM showing fewer false class positives.}
    \label{fig:SAP}
\vspace{-13pt}
\end{figure*}
%%%%%%%%%%%%%%%%%%%%%%%%%%%%%%%%%%%%%%%%%%%%%%%%%%%%%%%%%%

\vspace{3pt}
\noindent\textbf{Complementary Branch (\ourmodel{}).}
As described above, CAK branch has high class precision and low semantic sensitivity, while SAK branch has low class precision and high semantic sensitivity. Thus, we identify that the problem boils down to deriving an ideal object localization map which has high class precision \textit{and} high semantic sensitivity. This motivates the complementary learning of our framework where one branch carefully guides the beneficial, complementary knowledge to its counterpart branch. In particular, the CAK branch provides class-aware knowledge from its pseudo labels to the SAP of the SAK branch ($\mathcal{L}_{sap}$ in Fig.~\ref{fig:our_model}). Conversely, the SAK branch provides semantic-aware knowledge from its patch affinity map to the CAP of the CAK branch ($\mathcal{L}_{cap}$ in Fig.~\ref{fig:our_model}). 
In Sec.~\ref{sec:method3}, we show how the InfoNCE loss \cite{infonce} is carefully adapted for this process. 
Complementary knowledge guides each branch to minimize weaknesses and preserve strengths.

%%%%%%%%%%%%%%%%% 3.2. Branch Details  %%%%%%%%%%%%%%%%%%%%
\subsection{Branch Details}
\vspace{-6pt}
\label{sec:method2}
In this section, we provide details of each branch including the backbone architecture and various outputs. Both CAK and SAK branches take the same input image of size.
%%%%%%%%%%%%%%%%%%%%%%%%%%%%%%%%%%%%%%%%%%%%%%%%%%%%%%%%%%
\vspace{3px}

\noindent\textbf{CAK Branch.}
The CAMs are generated with the last downsampling layer to maintain resolution after passing through four layer blocks. We generate the following outputs using the feature map from the previous process (all found in Fig.~\ref{fig:our_model} top branch):
\begin{enumerate}[leftmargin=*]
    \item \textbf{CNN CAMs}: $f_{CAM}$ 
    takes the feature map to generate CNN CAMs ($\mathbf{M}_{cnn} \in \mathbb{R}^{N \times N \times C}$). This, after global average pooling, produces the image-level class scores for the classification loss $\mathcal{L}_{{cls}-{cnn}}$. $\mathbf{M}_{cnn}$ becomes the basis of the CNN-based object localization map.
    \item \textbf{Pseudo Labels}: 
    $\mathbf{M}_{cnn}$ are softmaxed on a class-wise basis to calculate the probability values for positive class. The class with the highest probability is selected using argmax to produce the $N \times N$ pseudo labels. 
    \item \textbf{Class-Aware Projection (CAP)}: Using the linear projection head $f_{proj}$, we generate CAP $\mathbf{v}_i^c \in \mathbb{R}^{128}$ for each CAM pixel $i \in \{1,2,\dots,N^2\}$. Thus, each $\mathbf{v}_i^c$ specifically entails the class-aware knowledge from the CNN CAMs. The semantic knowledge from SAK will refine these representations in Sec.~\ref{sec:method3}.
\end{enumerate}
%%%%%%%%%%%%%%%%%%%%%%%%%%%%%%%%%%%%%%%%%%%%%%%%%%%%%%%%%%
\vspace{3px}
\noindent\textbf{SAK Branch.}
The ViT model of our model follow its input convention ($N \times N$ patches with one class token).
The resulting $N^2$ tokens (i.e., $\mathbf{X} \in \mathbb{R}^{(N^2+1) \times D}$ where the embedding dimension $D$ is typically predefined) are transformed into three sets of vectors: query $\mathbf{Q}$, key $\mathbf{K}$, and value $\mathbf{V}$. Then, we calculate a new set of vectors based on the following attention function: $Attention(\mathbf{Q, K, V}) = softmax(\mathbf{QK}^T/\sqrt{D})\mathbf{V}$.
This common self-attention mechanism is the basis of ViT-based models where the attention map $\mathbf{A} = \mathbf{QK}^T$ of size $\mathbb{R}^{(1+N^2)\times (1+N^2)}$ is extracted (\textit{Attention Map} in Fig.~\ref{fig:our_model}). Now, we generate the following outputs from SAK branch:
\begin{enumerate}[leftmargin=*]
    \item \textbf{ViT CAMs}: The final tokens of ViT, excluding the class token size $\mathbb{R}^{N^2 \times D}$ (\textit{Patch Embedding} in Fig.~\ref{fig:our_model}) is fed to $f_{CAM}$ to generate the ViT CAMs  ($\mathbf{M}_{tran} \in \mathbb{R}^{N \times N \times C}$).
    \item \textbf{Semantic-Aware Projection (SAP)}: Using the linear projection head $f_{proj}$ in SAK branch, we generate SAP $\mathbf{v}_i^s \in \mathbb{R}^{128}$ for each patch, $i \in \{1,2,\dots,N^2\}$. Each $\mathbf{v}_i^s$ specifically contains the semantic-aware knowledge of the ViT tokens. The class-aware knowledge from CAK branch refines these tokens in Sec.~\ref{sec:method3}.
    \item \textbf{Patch Affinity}: A unique product of ViT is the patch affinity (\textit{Patch Affinity} in Fig.~\ref{fig:our_model}) derived from the attention map $\mathbf{A}$. 
    Specifically, the patch affinity is a $N^2 \times N^2$ dimensional attention map which is the average of the $L$ layers attention maps corresponding to the pairwise patches, namely, the following submatrix: $\mathbf{A}[1:(N^2+1), 1:(N^2+1)]$. 
\end{enumerate}
%%%%%%%%%%%%%%%%%%%%%%%%%%%%%%%%%%%%%%%%%%%%%%%%%%%%%%%%%%
\vspace{-8pt}
\subsection{Complementary Branch Losses}
\vspace{-5pt}
\label{sec:method3}
In this section, we describe the losses involved in \ourmodel{}. We first explain two losses necessary for basic functionality. Then, we discuss the complementary branch losses which explicitly provide class-aware and semantic-aware knowledge to the complementary branch in a ``one-way'' manner.
%%%%%%%%%%%%%%%%%%%%%%%%%%%%%%%%%%%%%%%%%%%%%%%%%%%%%%%%%% 
\vspace{-26pt}
\subsubsection{Class and CAM Losses}
\vspace{-7pt}
CNN CAMs and ViT CAMs make the image-level class predictions by minimizing $\mathcal{L}_{{cls}-{cnn}}$ and $\mathcal{L}_{{cls}-{tran}}$ softmax losses, which we combine as a single classification loss: $\mathcal{L}_{cls} = \mathcal{L}_{cls-cnn} + \mathcal{L}_{cls-tran}$.
Also, after a single epoch of training, we enforce a simple consistency between the CNN CAMs $\mathbf{M}_{cnn}$ and ViT CAMs $\mathbf{M}_{tran}$ via an L1 loss between them. In practice, since the localization maps are generated for the positive classes, an extra loss on the CAMs of positive classes ($\mathbf{M}^{+}_{cnn}$ and $\mathbf{M}^{+}_{tran}$) emphasizes their consistency: $\mathcal{L}_{cam} = ||{\mathbf{M}}_{cnn} - {\mathbf{M}}_{tran} ||_1 + || {\mathbf{M}}^{+}_{cnn} - {\mathbf{M}}^{+}_{tran} ||_1$.

\begin{figure*}[t!]
    \centering
    \includegraphics[width=0.9\textwidth]{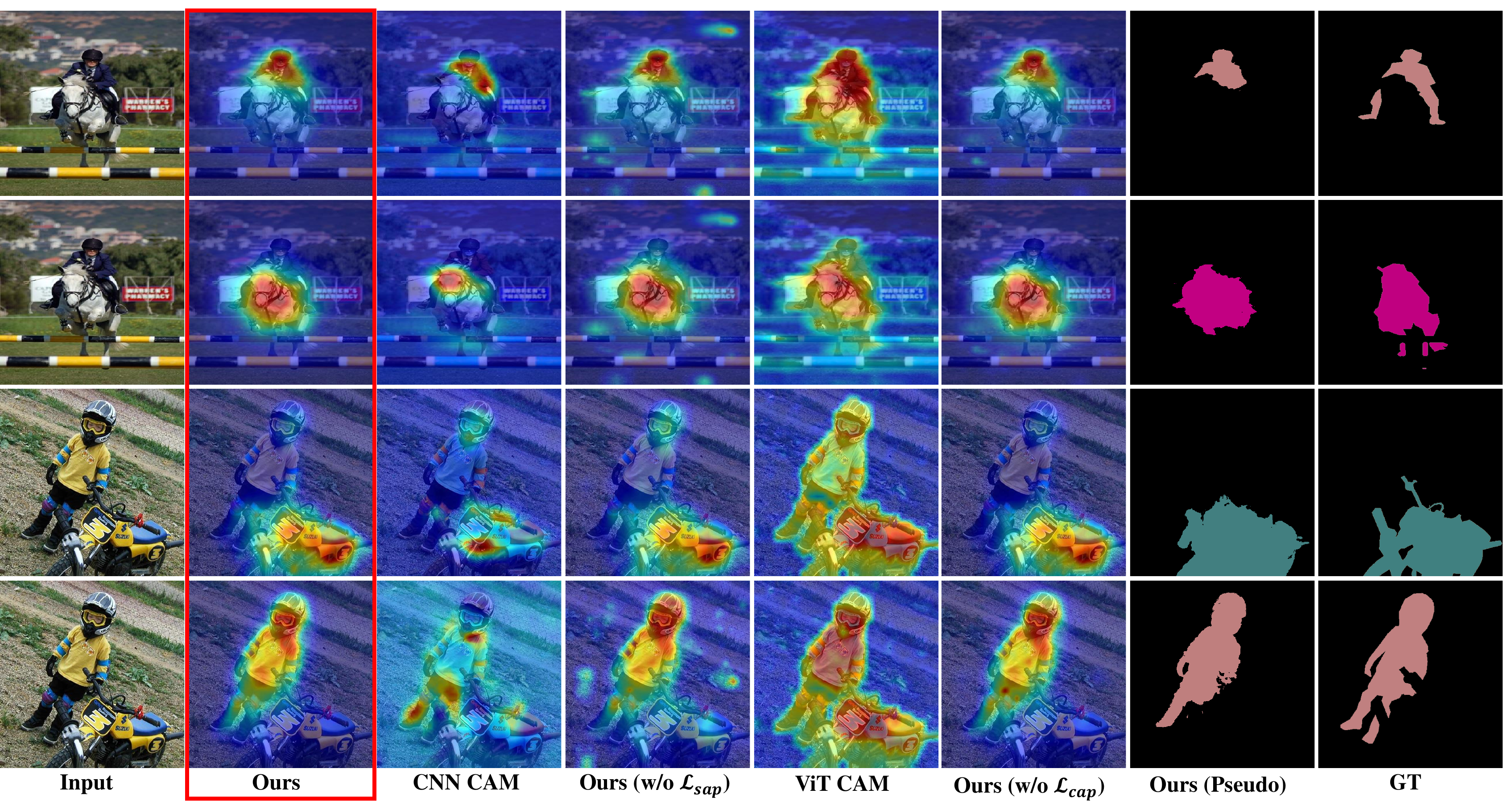}
\vspace{-8pt}
    \caption{Qualitative results. From left: (1) Input image, (2) Our result, (3) CNN CAM of our model, (4) Ours without $\mathcal{L}_{sap}$, (5) ViT CAM of our model, (6) Ours without $\mathcal{L}_{cap}$, (7) Our Pseudo mask for segmentation and (8) ground-truth segmentation label. We see that our results are able to differentiate between classes while finding their accurate object boundaries.
    }
    \label{fig:qual_result}
\vspace{-13pt}
\end{figure*}

\vspace{-15pt}
\subsubsection{Refining Class-Aware Projection}
\vspace{-7pt}
The key feature of CoBra is its complementary branch losses, injecting knowledge from one branch to another.
We first describe how CAP is refined using the semantic-aware knowledge from the patch affinity in SAK branch.
The patch affinity matrix $\mathbf{P} \in \mathbb{R}^{N^2 \times N^2}$ contains the semantic relations between all pairwise patches.
For instance, the  $i^{th}$ row $\mathbf{P}_i \in \mathbb{R}^{1 \times N^2}$ shows the semantic relationship between the $i^{th}$ patch to all $N^2$ patches.
Thus, $\mathbf{P}_i$ allows us to identify \textit{semantically} most similar patches as positive ($P^+$), least similar as negative ($P^-$), and neither as neutral patches with respect to the $i^{th}$ patch.
This is illustrated in the left block of Fig.~\ref{fig:SAP} showing how $\mathbf{P}_1$ (i.e., the first row of $\mathbf{P}$) chooses the patches semantically closest to the target patch (i.e., parts related to \texttt{horse}). 
This becomes an important cue to refine the CNN CAM which often excludes semantically relevant patches (thus low semantic sensitivity).
Once we assign a patch to be either positive, negative, or neutral, we refine CAP such that for each positive target patch ($\mathbf{v}_i^c$ for $i \in P^+$), (1) we \textit{pull} other positive patches ($\mathbf{v}_j^c$ for $j \in P^+$) towards $\mathbf{v}_i^c$ and (2) we \textit{push} negative patches ($\mathbf{v}_k^c$ for $k \in P^-$) away from $\mathbf{v}_i^c$.
Specifically, for each \textit{target} patch projection $\mathbf{v}_i^c$, we compute the following CAP loss inspired by InfoNCE \cite{infonce}:
\begin{equation}\label{CAPloss}\small
\mathcal{L}^{i}_{cap} =  \frac{1}{\left\vert P_{i}^{+} \right\vert} \sum_{j \in P_{i}^{+}} - \log\frac{\exp({\vv_i^c}{\vv_j^c}/{\tau})}{\exp({\vv_i^c}{\vv_j^c}/{\tau}) + \sum_{k \in {P_{i}^{-}}}\exp({\vv_i^c}{\vv_k^c}/{\tau})}
\end{equation}
where $\tau > 0$ is a temperature term. As seen in the top row of Fig.~\ref{fig:SAP}, CNN CAM has low semantic sensitivity, e.g., weakly localizing the \texttt{horse} regions other than the head.
The CAP loss refines the CAP representations such that  CNN CAM leverages the semantically positive patches (green circles) to further localize the remaining object parts.
Meaning, we observe that the semantic-aware knowledge of the SAK branch is provided as semantically relevant patches, and such \textit{high semantic sensitivity} information refines the CAP representations to reduce false negatives; in effect, improving the \textit{semantic sensitivity}.

\vspace{-15pt}
\subsubsection{Refining Semantic-Aware Projection} 
\vspace{-7pt}
We now describe how SAP is refined using the class-aware knowledge from the pseudo labels in CAK branch. 
Specifically, the pseudo label is an $N \times N$ matrix where each entry aligns to a class label from the CNN CAM location.
For instance, if a patch in CNN CAM has the strongest activation response to class \texttt{horse}, the corresponding patch entry of the pseudo label is assigned as \texttt{horse}.
This allows us to identify the patches corresponding to specific \textit{classes}.
In Fig.~\ref{fig:SAP}, the CNN CAM of a horse is used to identify positive patches $P^+$ (i.e., high activation score), negative patches $P^-$ (i.e., low activation score), and neutral patches (i.e., neither high nor low horse score).
% Once we assign a patch to be either positive, negative, or neutral, we refine SAP such that for each positive target patch ($\mathbf{v}_i^s$ for $i \in P^+$), (1) we \textit{pull} other positive patches ($\mathbf{v}_j^s$ for $j \in P^+$) towards $\mathbf{v}_i^s$ and (2) we \textit{push} negative patches ($\mathbf{v}_k^s$ for $k \in P^-$) away from $\mathbf{v}_i^s$.
Analogous to the CAP loss, we show the SAP loss which pulls the semantically similar SAP $\vv_j^s$ (green circles) to the target SAP $\vv_i^s$ (white circle) and pushes away dissimilar SAP $\vv_k^s$ (red circles) away from the target: 
\begin{equation}\label{SAPloss}\small
\mathcal{L}^i_{sap} =  \frac{1}{\left\vert P_{i}^{+} \right\vert} \sum_{j \in P_{i}^{+}} - \log\frac{\exp({\vv_i^s}{\vv_{j}^s}/{\tau})}{\exp({\mathbf{v}_i^s}{\mathbf{v}_j^s}/{\tau}) + \sum_{k \in {P_{i}^{-}}} \exp({\vv_i^s}{\vv_k^s}/{\tau})}
\end{equation}
where $\tau > 0$ is a temperature term. 
As shown in Fig.~\ref{fig:SAP} bottom row, SAK branch where for a positive target patch (white patch), SAP is updated to pull the positive (green circles) towards and push negative (red circles) away from the positive target SAP features. The resulting SAP refined CAM is also shown in Fig.~\ref{fig:SAP} where the false positive ViT CAM patches (i.e., patches in ViT CAM incorrectly localizing horse) are correctly adjusted to avoid localizing non-horse regions. In other words, we observe that the class-aware knowledge of CAK branch is provided as class-specific patches, and such \textit{high class precision} information refines the SAP representations to reduce false class positives; in effect, improving the \textit{class precision} of the ViT CAM. In Sec.~\ref{exp:ablation2}, we show how we chose $|P^+|$ and $|P^-|$.

\vspace{4px}
\noindent\textbf{Final Loss Function.}
We let $\mathcal{L}_{sap} = \sum_i\mathcal{L}_{sap}^i$ and $\mathcal{L}_{cap} = \sum_i \mathcal{L}_{cap}^i$ to combine the losses across all possible $i \in \{1,\dots,N^2\}$ patches.  Then, $\lambda_1 >0$ and $\lambda_2 > 0$, the full objective function is defined as:
\vspace{-3px}
\begin{equation}\label{eq3}\small
\mathcal{L}_{total} =\mathcal{L}_{cls} + \lambda_1 \mathcal{L}_{cam}+ \lambda_2 (\mathcal{L}_{sap} + \mathcal{L}_{cap} ).
\end{equation}
%%%%%%%%%%%%%%%%%%%%%%%%%%%%%%%%%%%%%%%%%%%%%%%%%%%%%%%%%%

%%%%%%%%%%%%%%%%%%%%%%%%%%%%%%%%%%%%%%%%%%%%%%%%%%%%%%%%%%
\subsection{Seed and Mask Generation}
\vspace{-5pt}
\label{sec:method4}
After training our model, we generate a \textit{seed} (i.e., CAM) for each image, which is then used to create a pseudo mask. This \textit{mask} serves as the final segmentation label, derived from image-level weak supervision. Note that each branch generates its own CAM; thus, we discuss the available options for the best mask generation.

\vspace{4px}
\noindent\textbf{CNN CAM.}
The CAM (i.e., $\mathbf{M}_{cnn}$) generated by the CAK Branch is made after global average pooling.
% is Class-Aware CNN-based object localization map. 
Although $\mathbf{M}_{cnn}$ has become much more semantically precise due to the complementary fusion with SAK, it still slightly lacks the preciseness of the ViT CAM (i.e., $\mathbf{M}_{tran}$). 

\vspace{4px}
\noindent\textbf{ViT CAM.} 
To generate CAM from SAK Branch, we use semantic-agnostic attention map~\cite{TS-CAM} $A_{obj} = A[0, 1:(1+N^2)] \in \mathbb{R}^{1 \times N^2}$ based on the attention map $A$ to generate the \textit{localization map} as ${\mathbf{M}_{tran}} = {\mathbf{M}_{tran}} \otimes {\mathbf{A}_{obj}}$
where $\otimes$ is the element-wise product. Fusing the CAK branch results in more accurate class-specific localization maps in $\mathbf{M}_{tran}$.

\vspace{4px}
\noindent\textbf{Fusion CAM.}
To overcome the intrinsic differences between CNN and ViT while utilizing better CAMs, we employ a simple CAM fusing mechanism for merging $\mathbf{M}_{cnn}$ and $\mathbf{M}_{tran}$. It takes advantage of both CAMs:
$\mathbf{M}_{fuse} = \max \left\{ (\mathbf{M}_{cnn}+\mathbf{M}_{tran})/{2}, \hspace{2pt} \mathbf{M}_{tran}\right\}$. This simple but effective merging mechanism balances the two complementary knowledge by taking a simple average of $\mathbf{M}_{cnn}$ and  $\mathbf{M}_{tran}$ and maximizs to explicitly capture both strong class and semantic knowledge. $\mathbf{M}_{fuse}$ is further used for generating the discretized mask via common post-processing using denseCRF. More details in Sec.~\ref{sssec:ablation}

\vspace{-8pt}

\section{Experiments}
\vspace{-5pt}
\label{sec:experiments}
\begin{table}[!t]
\def\arraystretch{0.9}% 
\setlength{\tabcolsep}{12pt}
\centering\small
\caption{\label{table:result1} Evaluation of initial seed and corresponding pseudo segmentation mask on PASCAL VOC 2012 training set in mIoU (\( \% \)).
\vspace{-8pt}}
\begin{tabular}{lccc}
% \hline
\Xhline{1pt}
\textbf{Method} & \textbf{Venue} & \textbf{Seed} & \textbf{Mask} \\ \hline
PSA \cite{psa}             & CVPR'18        & 48.0          & 61.0          \\
IRN \cite{irnet}             & CVPR'19        & 48.8          & 66.3          \\
Chang et al.\cite{chang}    & CVPR'20        & 50.9          & 63.4          \\
CDA \cite{cda}            & ICCV'21        & 55.4          & 63.4          \\
SEAM \cite{seam}           & CVPR'21        & 55.4          & 63.6          \\
AdvCAM \cite{advcam}          & CVPR'21        & 55.6          & 68.0          \\
CPN \cite{crossimage_pixel_contrast}             & ICCV'21        & 57.4          & 67.8          \\
MCTformer \cite{mctformer}      & CVPR'22        & 61.7          & 69.1          \\
ReCAM \cite{recam}         & CVPR'22        & 56.6          & 70.5          \\ 
AEFT \cite{AEFT}             & ECCV'22          & 56.0      & 71.0 \\
ACR \cite{ACR}             & CVPR'23          & 60.3      & 72.3 \\
ACR+ViT \cite{ACR}             & CVPR'23          & \textbf{65.5}      & 70.9 \\ \hline
% \rowcolor{gray!20}
Ours            &                &  62.3            & \textbf{73.5}              \\ 
\Xhline{1pt}
\end{tabular}
\vspace{-10pt}
\end{table}
\vspace{-10pt}

\vspace{13px}
% \subsection{Setup}
\vspace{-5pt}
% \vspace{3px}
\noindent\textbf{Dataset.}
We evaluate our method and other baselines on PASCAL VOC2012 \cite{everingham2010pascal} and MS COCO 2014. PASCAL VOC2012 contains 20 foreground classes and 1 background class, comprised of 1,464 training, 1,449 validation, and 1,456 test sets. In practice, following the widely used convention from other literature, an augmented set of 10,582 images is used as the training set. MS COCO comprised of 80K and 40K for training and validation, respectively. It contains 80 object classes and one background class for semantic segmentation.

\noindent\textbf{Evaluation Metrics.}
We compute the mean Intersection-over-Union (mIoU) to assess the segmentation performance on the validation (\textbf{val}) and \textbf{test} sets. In particular, the test set is evaluated on the PASCAL VOC online evaluation server.

\noindent\textbf{Implementation Details.}
We use ResNet152 \cite{resnet} following IRN \cite{irnet} and DeiT-S/16 \cite{deit} pre-trained on ImageNet for CNN and ViT respectively. The training set images are randomly resized and cropped to $244 \times 244$. Following the prior works on semantic segmentation, we use DeeplabV3+ \cite{chen2018encoder}, SegFormer \cite{xie2021segformer} based on ResNet101, MiT-B2, respectively. During the test, we use multi-scale and post-process with denseCRF. For all experiments, we used NVIDIA RTX A6000 GPU. Details are in the supplement.
\vspace{3pt}

\begin{table}[!t]
\caption{\label{table:main results} Semantic segmentation results on the validation (\textbf{Val}) and \textbf{Test} set of PASCAL VOC 2012 dataset. \textbf{Sup.} (Supervision): Image (I) and Saliency Map (S).}
\vspace{-5pt}
\resizebox{\columnwidth}{!}{%
\begin{tabular}{lccccc}
\Xhline{1pt}
\textbf{Method}        & \textbf{Venue}      & \textbf{Backbone}     & \textbf{Sup.} & \textbf{Val}           & \textbf{Test}          \\ \hline
EDAM          & CVPR'21    & ResNet101    & I+S  & 70.9          & 70.6          \\
L2G \cite{l2g}           & CVPR'21    & ResNet101    & I+S  & 72.1          & 71.7          \\
EPS \cite{eps}          & CVPR'21    & ResNet101    & I+S  & 71.0          & 71.8          \\ \hline
OOA* \cite{irnet}          & ICCV'19    & ResNet101     & I    & 65.2          & 66.4          \\
IRN \cite{irnet}          & CVPR'19    & ResNet50     & I    & 63.5          & 64.8          \\
SEAM \cite{seam}          & CVPR'20    & ResNet38     & I    & 64.5          & 65.7          \\
CONTA \cite{conta}          & NeruIPS'20 & WideResNet38 & I    & 66.1          & 66.7          \\
CDA \cite{cda}           & ICCV'21    & ResNet38     & I    & 66.1          & 66.8          \\
ECS-Net \cite{ecsnet}       & ICCV'21    & ResNet38     & I    & 66.6          & 67.6          \\
CPN \cite{cpn}          & ICCV'21    & WideResNet38 & I    & 67.8          & 68.5          \\
AdvCAM \cite{advcam}       & CVPR'21    & ResNet101    & I    & 68.1          & 68.0          \\
ReCAM \cite{recam}        & CVPR'22    & ResNet101    & I    & 68.5          & 68.4          \\
MCTformer \cite{mctformer}    & CVPR'22    & WideResNet38     & I    & 71.9          & 71.6          \\ 
AMN \cite{amn}    & CVPR'22    & ResNet101     & I    & 70.7          & 70.6          \\ 
AEFT \cite{AEFT}        & ECCV'22    & ResNet101    & I    & 70.0          & 71.3          \\
BECO \cite{Beco}        & CVPR'23    & ResNet101    & I    & 72.1         & 71.8          \\
OCR+MCTformer \cite{ocr}        & CVPR'23    & WideResNet38    & I    & 72.7         & 72.0          \\
ACR \cite{ACR}        & CVPR'23    & WideResNet38    & I    & 71.9         & 71.9          \\
ACR+ViT \cite{ACR}        & CVPR'23    & WideResNet38    & I    & 72.4         & 72.4          \\ \hline
% \rowcolor{gray!20}
\textbf{Ours} &            &    ResNet101          & I    & \textbf{74.0} & \textbf{73.9} \\ \hline
AFA \cite{spatialBCE}        & CVPR'23    & MiT-B1    & I    & 66.0         & 66.3          \\
BECO \cite{Beco}        & CVPR'23    & MiT-B2    & I    & 73.7         &  73.5          \\ \hline
% \rowcolor{gray!20}
\textbf{Ours} &            &    MiT-B2          & I    & \textbf{74.3} & \textbf{74.2} \\ 
\Xhline{1pt}
\end{tabular}%
}
\vspace{-5pt}
\end{table}
\vspace{-9pt}

%%%%%%%%%%%%%%%%%%%%%%%%%%%%%%%%% Loss function %%%%%%%%%%%%%%%%%%%%%%%%%%%%%%%%%%%%%%

\begin{table} % [!t]
\def\arraystretch{0.9}% 
\centering\small
\caption{\label{table:loss} Seed mIoU (\%) of varying loss function combinations.}
\vspace{-5pt}
\setlength{\tabcolsep}{4pt}
\begin{tabular}{cccccc}
\Xhline{1pt}
CAK branch & SAK branch & $\mathcal{L}_{cam}$ & $\mathcal{L}_{cap}$ & $\mathcal{L}_{sap}$ & mIoU (\%)     \\ \hline
$\checkmark$        &                     &                     &                      &                    & 48.8              \\
                    & $\checkmark$        &                    &                     &                      & 41.3         \\
$\checkmark$        & $\checkmark$        & $\checkmark$        &                     &                     & 53.2              \\
$\checkmark$        & $\checkmark$        &                     & $\checkmark$        & $\checkmark$        & 56.4         \\
$\checkmark$        & $\checkmark$        & $\checkmark$        & $\checkmark$        &                     & 59.2          \\
$\checkmark$        & $\checkmark$        & $\checkmark$        &                     & $\checkmark$        & 61.2          \\
$\checkmark$        & $\checkmark$        & $\checkmark$        & $\checkmark$        & $\checkmark$        & \textbf{62.3} \\
\cline{2-3}
\Xhline{1pt}
\end{tabular}
\vspace{-8pt}
\end{table}
%%%%%%%%%%%%%%%%%%%%%%%%%%%%%%%%%%%%%%%%%%%%%%%%%%%

\subsection{Evaluations}
\vspace{-7pt}
\noindent\textbf{Seed and Mask.}
Our work follows the family of existing steps which generates object localization maps (seed) and produces pseudo semantic segmentation ground-truth labels (mask) with IRN \cite{irnet}. Thus, we first evaluate the seed and mask quality by computing the mIoU on the ground-truth segmentation labels. In Table~\ref{table:result1}, \ourmodel{} achieves the second best seed (\textbf{62.3\%}) without crf and the best mask (\textbf{73.5\%}) compared to existing state-of-the-art methods.
Even without using CRF, our mask achieves \textbf{72.5}\% mIoU which still surpasses others using CRF. See the Supplement.

\noindent\textbf{Semantic Segmentation.}
Generated masks are used as pseudo labels for training a segmentation network, which is then evaluated on performance on the PASCAL VOC 2012 val and test datasets. As shown in Table~\ref{table:main results}, our model \ourmodel{} achieves the state-of-the-art results of \textbf{74.0\%} and \textbf{73.9\%} on val and test sets respectively for ResNet101 backbone and \textbf{74.3\%} and \textbf{74.2\%} for MiT-B2 backbone. The detailed results on MSCOCO are provided in the Supplement.

\subsection{Ablation Studies} \label{sssec:ablation}
\vspace{-5pt}

\noindent\textbf{Effect of Cross Complementary Branch.}
The importance of our proposed dual-branch scheme is highlighted by evaluating the model with only one branch at a time. Specifically, when our model operates with a single branch and without cross-complementary losses, the CAK and SAK branches correspond closely to IRNet \cite{irnet} and TS-CAM \cite{TS-CAM}, respectively. We make minimal changes to properly produce the localization maps and assess their quality. In Table~\ref{table:loss}, the single branch setups exhibit lower mIoU compared to our dual branch setup, as shown in Fig.~\ref{fig:qual_result}.
% In Table~\ref{table:loss}, we observe the single branch setups show relatively low mIoU compared to our dual branch setup and qualitative results in Fig.~\ref{fig:qual_result}.
% Qualitative results in Fig.~\ref{fig:qual_result}. 

\begin{table}[!t] \small
\def\arraystretch{0.9}% 
\caption{\label{table:topk} Seed mIoU (\%) using various combinations of $k_{sap}^-$ and $k_{cap}^+$ while fixing $k_{sap}^+=20$ and $k_{cap}^-=20$.
\vspace{-5pt}}
\resizebox{\columnwidth}{!}{%
\setlength{\tabcolsep}{14pt}
\begin{tabular}{c|cccc}
\Xhline{1pt} 
\diagbox[width=7em]{$k_{sap}^-$}{$k_{cap}^+$}& 5           & 10 & 15 & 20 \\ \hline
5               & \textbf{62.3} & 61.9 & 59.1 & 56.8 \\
10              & 61.3          & 60.0 & 58.4 & 55.7 \\
15              & 59.1         & 58.3 & 56.3 & 54.6 \\
20              & 58.9         & 57.2 & 55.4    & 53.8     \\
\Xhline{1pt}
\end{tabular}%
} 
\vspace{-10pt}
\end{table}

\begin{table}[!t] \small
\def\arraystretch{0.95}% 
\caption{\label{table:generating} Mask mIoU (\%) under various sources.}
\vspace{-5pt}
\centering
\setlength{\tabcolsep}{15pt}
\begin{tabular}{cc}
\Xhline{1pt}
\textbf{Source for Mask} & \textbf{Mask mIoU (\%)} \\ \hline
$\mathbf{M}_{cnn}$      & 71.6       \\
$\mathbf{M}_{tran}$     & 71.5    \\
$(\mathbf{M}_{cnn} + \mathbf{M}_{tran})/2$     & 72.2      \\
$\max \left\{ \mathbf{M}_{cnn}, \mathbf{M}_{tran}\right\}$      & 72.6       \\ \hline
$\mathbf{M}_{fuse}$     & \textbf{73.5}      \\ 
\Xhline{1pt}
\end{tabular}%
%}
\vspace{-10pt}
\end{table}

\noindent\textbf{Loss Functions.}
We perform a detailed examination of the impact of various loss functions. Table~\ref{table:loss} shows that each of $\mathcal{L}_{sap}$ and $\mathcal{L}_{cap}$ brings noticeable improvements while using all three losses is the best. This further implies that the complementary class and semantic knowledge with $\mathcal{L}_{cap}$ and $\mathcal{L}_{sap}$ bring significant benefits while retaining the similarity between the CNN and ViT CAMs via $\mathcal{L}_{cam}$.

\vspace{2px}
\noindent\textbf{Choosing Positive and Negative Patches.}\label{exp:ablation2} Recall that the SAP loss (Eq.~\eqref{SAPloss}) in the SAK branch learns from the complementary CAK pseudo label patches. Specifically, $P_{sap}^+$ is the top $k_{sap}^+$ scoring positive pseudo label patches from CAK branch, and $P_{sap}^-$ is the bottom $k_{sap}^-$ scoring negative pseudo label patches from CAK branch. Similarly, in the CAP loss (Eq.~\eqref{CAPloss}) in the CAK branch learns from the complementary SAK patch affinity map. That is, $P_{cap}^+$ is the top $k_{cap}^+$ scoring positive pseudo label patch affinity map, and $P_{cap}^-$ is the bottom $k_{cap}^-$ scoring negative patch affinity map.

In this study, we analyze how $k_{sap}^+$, $k_{sap}^-$, $k_{cap}^+$, and $k_{cap}^-$ affect the overall outcome. As we have observed before, the patches with positive pseudo labels from the CAK branch are highly precise. In other words, the positive pseudo labels from CAK are \textit{highly likely to be true positives}. Thus, we explicitly set $k_{sap}^+$ to be 20 (approximately 10\% of the total patches) to maximally incorporate the confident knowledge about true positives from the CAK branch into the SAP loss.

Similarly, the patches with negative patch affinity map from the SAK branch are also highly precise. In other words, the negative patch affinity map from SAK branch is \textit{highly likely to be true negatives}. Thus, we also set $k_{cap}^-$ to be 20 to maximize the confident knowledge about true negatives from the SAK branch into the CAP loss.

Table~\ref{table:topk}, we show the seed mIoU of various combinations of $k_{sap}^-$ and $k_{cap}^+$ for fixed $k_{cap}^+=20$ and $k_{cap}^-=20$. We observe that $k_{sap}^-=5$ and $k_{cap}^+=5$ showed the best result, implying that leveraging a relatively small number of competent patches brings robustness to our model.

\begin{figure}[t!]
    \centering
    \includegraphics[width=1\columnwidth]{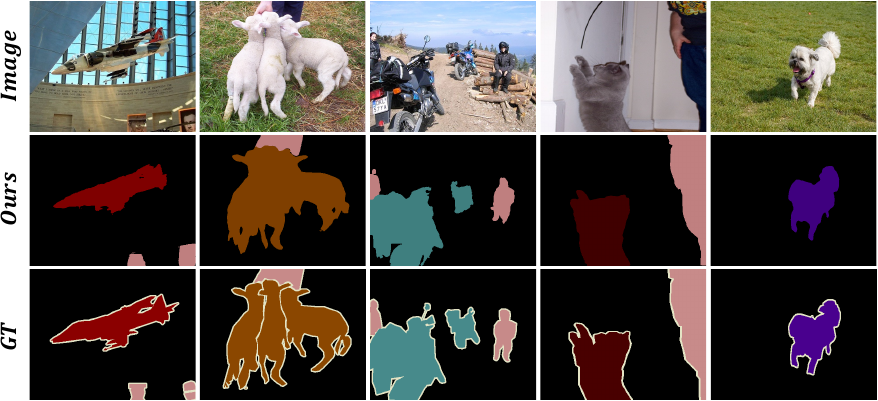}
    \vspace{-20pt}
    \caption{Qualitative \textit{seg} results on the PASCAL VOC \textit{val} set.}
    \label{fig:CAP}
    \vspace{-18pt}
\end{figure}

\vspace{-5pt}

\vspace{5px}
\noindent\textbf{Ensuring Branch Benefits.}
We note that it is crucial to ensure that the branches benefit each other while suppressing the drawbacks. Empirically, $\mathcal{L}_{cam}$ (53.2\%) which mutually influences both branches outperforms each SAK (48.8\%) and CAK (41.3\%) results, indicating that the benefits outweigh the potential flow of drawbacks. Also, $\mathcal{L}_{cap}$ \& $\mathcal{L}_{sap}$ mechanically ensure the ``one-way'' flow of benefits from one branch to the other by selectively updating only the receiving branch. Together with using the top $k$ competent patches, we quantitatively check how these losses robustly bring improvements as in Table~\ref{table:loss}.

\noindent\textbf{Seed Fusion for Mask Generation.}
Our model, by construction, provides two complementarily useful seeds, namely, $\mathbf{M}_{cnn}$ and $\mathbf{M}_{tran}$ which may be carefully fused for robust outcomes. In particular, as we discussed in Sec.~\ref{sec:method4}, we have experimented with various combinations of the seeds as shown in Table~\ref{table:generating}. It shows combining $\mathbf{M}_{cnn}$ and $\mathbf{M}_{tran}$ yields better results than using them individually. 
Considering the unique characteristics of each CAM, we discovered that the best results were obtained when they were fused as in $\mathbf{M}_{fuse}$. More details in the supplement.

\vspace{-5px}

\vspace{-4pt}

\section{Conclusion}
\vspace{-7pt}
\label{sec:conclusions}
In this work, we proposed a novel dual-branch framework for fusing complementary knowledge from CNN and vision transformer for WSSS. The quantitative results show state-of-the-art performance on the PASCAL VOC 2012 datasets and MS COCO 2014 datasets, while qualitatively demonstrating the significance of properly exchanging class- and semantic-aware knowledge via Class-Aware Projection and Semantic-Aware Projection. We particularly hope this work brings new insights toward considering both CNN and vision transformers as two equally crucial, complementary counterparts.

{\small
\bibliographystyle{ieeenat_fullname}
\bibliography{main}
}

\clearpage 
\onecolumn 
\section*{Supplementary Material} 
This supplementary material provides additional details on our experiments, including additional results for \textbf{MS-COCO 2014 dataset} (Section.~\ref{subsec: additional study}), Additional visualization and analyses of the results (Section.~\ref{subsec: vis}). \textbf{Ablation study} for k values and embedding size are also provided (Section.~\ref{subsec: ablation}). Further details and examples of our analysis on fusing ViT and CNN cam while generating masks (Section.~\ref{subsec: mask}). The experimental setup (Section.~\ref{sec: 1}) and the \textbf{discussion} on the Methodology for Determining Thresholds (Section.~\ref{sec: thres}) are both included. We also have described our results in an easily accessible manner on our project page. The link to the \textbf{project page} is as follows: \url{https://micv-yonsei.github.io/cobra2024}.

\section{Additional Study for another dataset}
\label{subsec: additional study}
In this section, we will discuss the segmentation results for the MS-COCO 2014 dataset.
\subsection{MS-COCO 2014}

\begingroup
\setlength{\tabcolsep}{10pt} 
\renewcommand{\arraystretch}{1.0}
\begin{table}[!h]
\caption{Semgentation mIoU results(\%) on MS-COCO 2014 \textit{val} dataset.}
    \centering
    \begin{tabular}{l c c c}
        \hline
        Method & Venue & Backbone & val \\
        \hline
        IRNet~\cite{irnet}& CVPR'19 & ResNet50 & 41.4 \\
        SEAM~\cite{seam}& CVPR'20 & WideResNet38 & 31.9  \\
        OC-CSE~\cite{oc-cse} & ICCV'21 & WideResNet38 & 36.4  \\
        PMM~\cite{pmm} & ICCV'21 & WideResNet38 & 36.7  \\
        MCTformer~\cite{mctformer} & CVPR'22 & WideResNet38 & 42.0 \\
        URN~\cite{urn} & CVPR'22 & WideResNet38 & 40.7  \\
        SIPE~\cite{SIPE} & CVPR'22 & WideResNet38 & 43.6  \\
        Spatial-BCE~\cite{spatialBCE} & ECCV'22 & VGG16 & 35.2  \\
        ACR~\cite{ACR} & CVPR'23 & WideResNet38 & 45.3  \\
        BECO~\cite{Beco} & CVPR'23 & ResNet101 & 45.1  \\ \hline
        Ours &  & ResNet101 & \textbf{45.5} \\ \hline
    \end{tabular}
    \label{table:mscoco}
\end{table}
\endgroup

To validate the consistent performance of our methodology across diverse data types, we executed segmentation experiments on the MS-COCO 2014 dataset as seen in Table.~\ref{table:mscoco}. For these experiments, we employed ResNet101, following the previous studies~\cite{Beco}. Our results show impressive outcomes on the MS-COCO 2014 dataset. This not only confirms the consistency of our approach across different datasets but also indicates progress in the field. Specifically, our method achieves a marked improvement, with a gain of state-of-the-art MS-COCO 2014 dataset.

\section{Additional Visualization and Analyses of the Results}
\label{subsec: vis}
In this section, we provide additional visualizations and analyses of the results. In Section.~\ref{subsec: segresults}, we present further visualizations of the segmentation results. Section.~\ref{subsec: crf} is dedicated to analyzing the impact of CRF, which is essential in the creation of pseudo labels. Lastly, in Section.~\ref{subsec: class}, we document the mean Intersection over Union (mIoU) values for each class, focusing on our results, seeds, masks, and segmentation.

\subsection{Qualitative Segmentation Results}
\label{subsec: segresults}
\begin{figure}[t!]
    \centering
    \includegraphics[width=1\columnwidth]{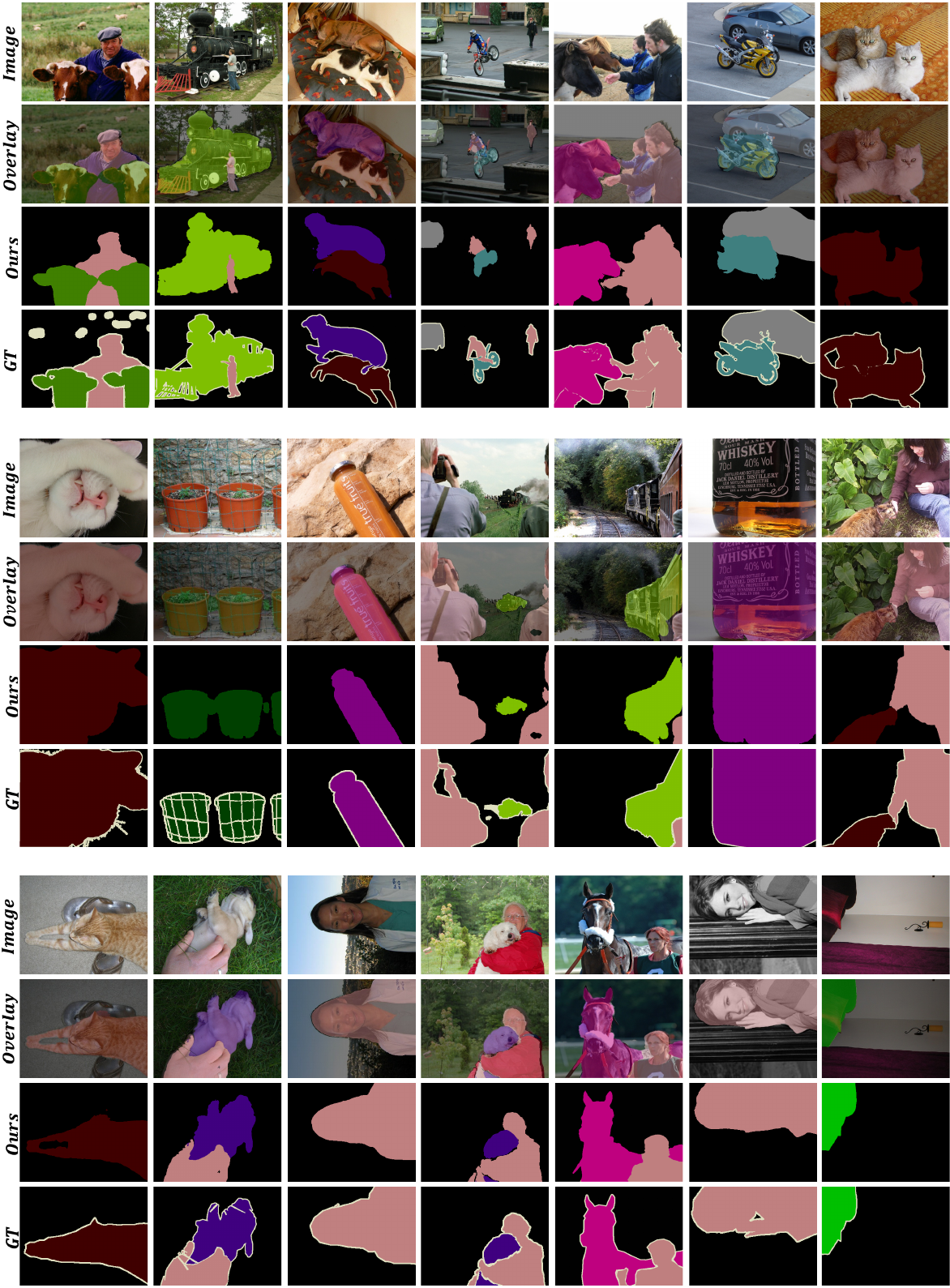}
    \caption{Additional qualitative segmentation results on the PASCAL VOC \textit{val} set, including the overlay of the segmentation results on images.}
    \label{fig:sup_fig0}
\end{figure}

\begin{figure}[t!]
    \centering
    \includegraphics[width=1\columnwidth]{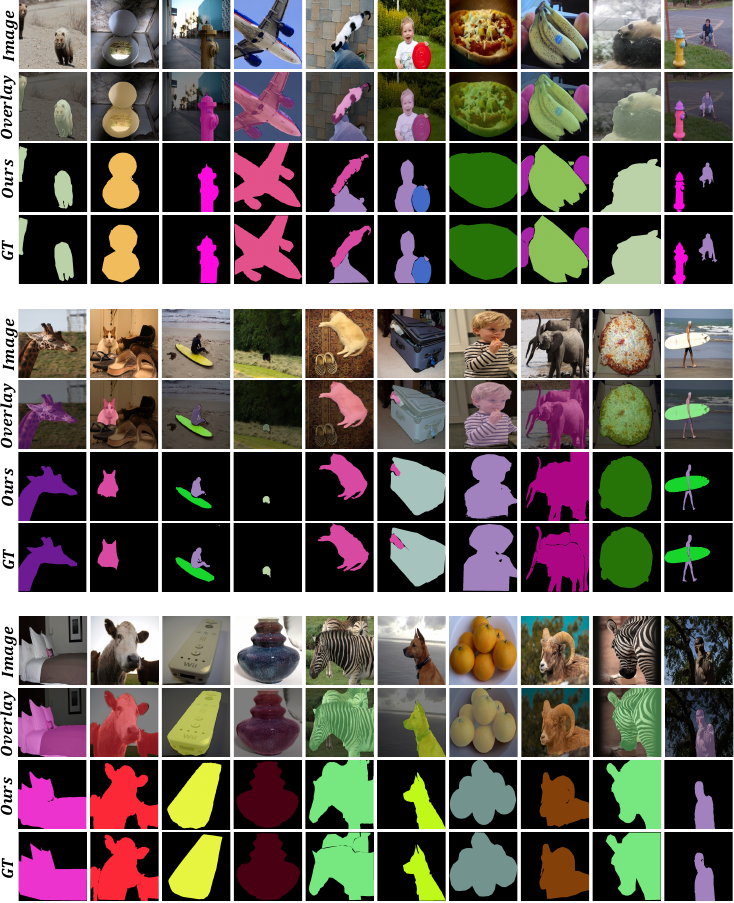}
    \caption{Additional qualitative segmentation results on the MS-COCO \textit{val} set, including the overlay of the segmentation results on images.}
    \label{fig:sup_fig1}
\end{figure}

In Fig.~\ref{fig:sup_fig0}, we provide more samples on the PASCAL VOC 2012 \textit{val} dataset. In Table.~\ref{table:sup_tab2}, mIoU values are presented for each class in the Pascal VOC 2012 dataset. More visualizations can be found on the following page. 
Additionally, in Fig.~\ref{fig:sup_fig1}, we provide more samples on the MS-COCO \textit{val} dataset. In Table.~\ref{table:sup_tab3}, mIoU and accuracy values are presented for each class in the MS-COCO 2014 dataset.

\newpage
\clearpage

\begin{table}[h]
\caption{\label{table:sup_tab2} mIoU(\%) scores for seed, mask, and segmentation categorized by class on Pascal VOC2012 dataset.}
\resizebox{\columnwidth}{!}{
\begin{tabular}{c|c|ccccccccccccccccccccc|c}
\hline
\multicolumn{2}{c|}{\diagbox[width=14em]{\textbf{model}}{\textbf{class}}} &
  background & 
  airplane &
  bike &
  bird &
  boat &
  bottle &
  bus &
  car &
  cat &
  chair &
  cow &
  table &
  dog &
  horse &
  mbk &
  person &
  plant &
  sheep &
  sofa &
  train &
  tv &
  mIoU(\%) \\ \hline
\multicolumn{2}{c|}{seed(\textit{train})} &
  85.26 &
  59.90 &
  37.03 &
  68.79 &
  48.30 &
  58.55 &
  77.13 &
  68.00 &
  70.89 &
  29.19 &
  72.04 &
  55.53 &
  75.68 &
  70.55 &
  70.74 &
  63.74 &
  56.70 &
  76.71 &
  56.86 &
  64.81 &
  41.41 &
 62.28 \\   
\multicolumn{2}{c|}{mask(\textit{train})} &
  90.70 &
  79.83 &
  42.42 &
  83.74 &
  68.26 &
  71.63 &
  86.42 &
  79.88 &
  87.39 &
  32.75 &
  85.23 &
  62.48 &
  86.86 &
  82.84 &
  78.19 &
  76.83 &
  64.50 &
  88.33 &
  75.03 &
  73.46 &
  45.87 &
  73.45 \\
\hline
\multirow{2}{*}{\textsc{ResNet101}} &
seg (\textit{val}) &
92.17 &
85.53 &
45.47 &
86.9 &
72.64 &
77.89 &
89.67 &
84.26 &
86.33 &
37.94 &
80.32 &
47.49 &
84.79 &
83.11 &
80.87 &
83.72 &
61.13 &
82.77 &
52.61 &
82.36 &
56.25 &
74.01 \\ 
&
seg (\textit{test}) &
94.97 &
90.03 &
37.24 &
85.43 &
64.3 &
71.84 &
93.02 &
86.33 &
91.82 &
25.2 &
83.42 &
59.62 &
87.59 &
85.53 &
86.48 &
84.38 &
64.14 &
85.33 &
57.84 &
73.24 &
44.97 &
73.94  \\ \hline
\multirow{2}{*}{\textsc{MiT-B2}} &
seg (\textit{val}) &
94.90 &
87.39 &
36.08 &
89.63 &
72.17 &
74.55 &
92.12 &
84.27 &
92.47 &
28.45 &
92.54 &
45.57 &
88.29 &
87.44 &
80.74 &
83.71 &
64.38 &
93.13 &
48.92 &
78.17 &
45.37 &
74.30 \\ 
&
seg (\textit{test}) &
95.79 &
89.88 &
35.74 &
82.82 &
65.33 &
70.51 &
90.87 &
85.75 &
94.08 &
24.92 &
87.13 &
60.80 &
89.87 &
86.52 &
86.50 &
84.17 &
65.83 &
87.55 &
55.58 &
71.50 &
47.08 &
74.20 \\
\hline
\end{tabular}
}
\end{table}

\begin{table}[h]
\centering
\caption{\label{table:sup_tab3} mIoU and accuracy for segmentation categorized by class on MS-COCO 2014 dataset}
\begin{tabular}{l|cc|l|cc}
\hline{}
\textbf{Class}        & \textbf{mIoU}  & \textbf{mAcc} & \textbf{Class}        & \textbf{mIoU}  & \textbf{mAcc}  \\ \hline{}
background            & 83.18         & 87.04  & person        & 65.23         & 72.44         \\ 
bicycle             & 51.23         & 59.2  & car             & 51.1         & 59.21        \\ 
motorcycle               & 65.97         & 72.21  & airplane            & 66.68         & 72.13        \\
bus                   & 66.61 & 72.49            & train           & 66.55 & 72.2         \\
truck          & 48.95 & 57.35 & boat           & 46.69 & 55.3   \\
traffic light  & 53.31 & 60.57 & fire hydrant   & 74.61 & 79.51 \\
stop sign      & 77.83 & 82.69 & parking meter  & 66.58 & 72.66 \\
bench          & 37.33 & 45.17 & bird           & 65.16 & 72.16 \\
cat            & 79.27 & 83.71 & dog            & 69.95 & 75.22 \\
horse          & 67.68 & 73.0  & sheep          & 71.67 & 76.6  \\
cow            & 69.98 & 75.03 & elephant       & 82.45 & 86.55 \\
bear           & 81.05 & 85.34 & zebra          & 83.72 & 87.34 \\
giraffe        & 78.07 & 82.69 & backpack       & 17.74 & 22.21 \\
umbrella       & 64.55 & 72.31 & handbag        & 49.09 & 57.35 \\
tie            & 29.62 & 36.7  & suitcase       & 50.76 & 58.99 \\
frisbee        & 48.08 & 56.63 & skis           & 15.8  & 20.03 \\
snowboard      & 35.96 & 43.74 & sports ball    & 11.54 & 14.74 \\
kite           & 30.78 & 38.03 & baseball bat   & 1.17  & 1.51  \\
baseball glove & 0.91  & 1.18  & skateboard     & 9.84  & 12.61 \\
surfboard      & 50.53 & 58.88 & tennis racket  & 1.81  & 2.33  \\
bottle         & 35.69 & 43.54 & wine glass     & 31.49 & 38.81 \\
cup            & 33.55 & 41.14 & fork           & 16.96 & 21.29 \\
knife          & 14.9  & 18.99 & spoon          & 8.38  & 10.76 \\
bowl           & 26.65 & 33.28 & banana         & 65.55 & 71.96 \\
apple          & 48.62 & 57.12 & sandwich       & 44.28 & 53.02 \\
orange         & 65.48 & 72.31 & broccoli       & 54.53 & 61.79 \\
carrot         & 44.53 & 53.17 & hot dog        & 56.24 & 63.55 \\
pizza          & 64.9  & 72.29 & donut          & 58.77 & 66.02 \\
cake           & 51.59 & 59.28 & chair          & 27.0  & 33.54 \\
couch          & 39.52 & 47.57 & potted plant   & 16.52 & 20.79 \\
bed            & 52.22 & 59.84 & dining table   & 15.88 & 20.08 \\
toilet         & 65.52 & 72.13 & tv             & 51.66 & 59.53 \\
laptop         & 53.05 & 60.62 & mouse          & 16.44 & 20.74 \\
remote         & 46.34 & 55.04 & keyboard       & 53.28 & 60.71 \\
cell phone     & 57.26 & 64.51 & microwave      & 42.59 & 51.13 \\
oven           & 34.46 & 42.15 & toaster        & 0.0   & 0.0   \\
sink           & 32.52 & 39.98 & refrigerator   & 44.57 & 53.07 \\
book           & 36.75 & 44.59 & clock          & 47.53 & 56.15 \\
vase           & 26.82 & 33.41 & scissors       & 37.37 & 45.1  \\
teddy bear     & 64.68 & 72.25 & hair drier     & 0.0   & 0.0   \\ \cline{4-6}
toothbrush     & 15.11 & 19.21 & \textbf{Total}&\textbf{45.53} & \textbf{51.67} \\ \hline
\end{tabular}
\label{tab:sup_tab3}
\end{table}

\subsection{Comparison of CRF performance before creating pseudo labels}
\label{subsec: crf}
To more accurately assess the performance of our mask, we executed a separate experiment without the post-processing step of CRF.
Interestingly, as can be seen in Table.~\ref{table:result2}, the results of our mask, even without CRF, proved to be superior to existing methods that include CRF.

\begingroup
\setlength{\tabcolsep}{10pt} 
\renewcommand{\arraystretch}{1.0} 
\begin{table}[!h]
\caption{\label{table:result2} Mask mIoU values on the PASCAL VOC 2012 \textit{train} Dataset with and without the use of CRF.}
    \centering
    \begin{tabular}{cccc}
    \hline
    Method & Venue   & Mask w/o CRF     & Mask w/ CRF    \\ \hline
    SEAM~\cite{seam}  & CVPR'21 & 55.4          & 56.8          \\
    AdvCAM~\cite{advcam} & CVPR'21 & 55.6          & 62.1          \\
    AEFT~\cite{AEFT}   & ECCV'22 & 56.0          & 71.0          \\  \hline
    \textbf{Ours}   &         & \textbf{71.9} & \textbf{73.5} \\ 
    \hline
    \end{tabular}
\end{table}
\endgroup

\subsection{Performance scores for seed, mask, and segmentation categorized by class}
\label{subsec: class}
In this section, we have segmented and analyzed the performance of our model CoBra in terms of seed, mask, and segmentation (seg) values, broken down by class, and presented these as mIoU values. Table.~\ref{table:sup_tab2} presents the mIoU values for each class on the Pascal VOC 2012 dataset. The seed and mask in Table.~\ref{table:sup_tab2} are evaluated on the PASCAL VOC \textit{train} dataset. Meanwhile, Table.~\ref{table:sup_tab3} details the mIoU and accuracy values for each class based on the segmentation results obtained from the MS-COCO 2014 dataset.

\section{Additional Ablation Study}
\label{subsec: ablation}
In this section, we provide additional ablation analysis on the $k$ values~(Section. \ref{subsec: 2}) and exploration of embedding sizes (Section.~\ref{subsec: ablembed}).

\subsection{Ablation Study for $k$ values}
\label{subsec: 2}
In Table.~\ref{table:sup_tab1}, we performed ablation on $k^{+}_{sap}$ and $k^{-}_{cap}$, which are not demonstrated in the main paper. 
This proves our hypothesis that patches with positive labels from the CAK branch are highly precise, and similarly, the patches with negative patches from the affinity map derived from SAK branches are also highly precise. The max value was set to $20$, as it was observed that the ratio of true positives exists at about 10\% across all patches. Therefore, we concluded that the most reliable ratio stands at $10$\%. With a total of $196$ patches, each of size $14 \times 14$, we identified the top $10$\% - which amounts to $20$ patches - as the most trustworthy group for our study. Table.~\ref{table:sup_tab1}, we show the seed mIoU of various combinations of $k^{+}_{sap}$ and $k^{-}_{cap}$ for fixed $k^{-}_{sap} =5$ and $k^{+}_{cap} =5$. We observe that $k^{+}_{sap} = 20$ and $k^{-}_{cap} =20$ showed the best result.

\begin{table}[h!]\small
\def\arraystretch{1.0}% 
\centering
\caption{\label{table:sup_tab1} Seed mIoU (\%) using various combinations of $k_{sap}^+$ and $k_{cap}^-$ while fixing $k_{sap}^-=5$ and $k_{cap}^+=5$.}
\setlength{\tabcolsep}{15pt}
\begin{tabular}{c|cccc}
\hline
\diagbox[width=7em]{$k_{sap}^+$}{$k_{cap}^-$}& 5           & 10 & 15 & 20 \\ \hline
5               & 54.1 & 54.8 & 55.1 & 56.3 \\
10              & 56.4 & 57.0 & 57.4 & 58.5 \\
15              & 57.8 & 57.4 & 59.9 & 60.5 \\
20              & 58.1 & 58.6 & 60.1  & \textbf{62.3}     \\
\hline
\end{tabular}%
\end{table}

\subsection{Ablation Study for embedding size}
\label{subsec: ablembed}
In determining the optimal embed size for our model, we present a series of comparative validations across a range of sizes in Table.~\ref{table:table2_embed}. Although different embed sizes produced notable results, we found that the size of $256$, which is currently utilized by our CoBra model, consistently achieved the highest values.

\newpage
\clearpage

\begin{table}[h!]\small
\def\arraystretch{1.0}% 
\centering
\caption{\label{table:table2_embed} mIoU values(\%) for each seed based on embedding size.}
\setlength{\tabcolsep}{15pt}
\begin{tabular}{c|ccc}
\hline
Embedding size & 128  & 256           & 384  \\ \hline
seed mIoU(\%)           & 61.9 & \textbf{62.3} & 60.4 \\ \hline
\end{tabular}
\end{table}

\subsection{Ablation Study on Lambda Values Applied to the Loss Function}
\label{subsec: lambda}
The overall loss function is defined as:
\begin{equation}\label{sup_eq1}
\mathcal{L}_{total} =\mathcal{L}_{cls} + \lambda_1 \mathcal{L}_{cam}+ \lambda_2 (\mathcal{L}_{sap} + \mathcal{L}_{cap} ).
\end{equation}
After performing several tests to determine the optimal loss weights for 
$\mathcal{L}_{cam}$, $\mathcal{L}_{sap}$ and $\mathcal{L}_{cap}$ with various lambda values, it was found that the most favorable outcomes were achieved with $\lambda_1$ at 0.1 and $\lambda_2$ at 0.1. Moreover, it became apparent that our model's performance remained consistently high, relatively unaffected by the variations in lambda values.

\begin{table}[h] 
\def\arraystretch{1.0}% 
\centering
\caption{\label{table:sup_tab11} Seed mIoU (\%) using various $\lambda$ while calculating $\mathcal{L}_{SAP}$ and $\mathcal{L}_{CAP}$.}
\setlength{\tabcolsep}{20pt}
\begin{tabular}{cc|c}
\hline
$\lambda$ value of $\mathcal{L}_{SAP}$ & $\lambda$ value of $\mathcal{L}_{CAP}$ & mIoU(\%)      \\ \hline
0.05    & 0.05    & 59.5          \\
0.05    & 0.1     & 60.6              \\
0.1     & 0.05    & 59.4              \\ %\rowcolor{headercolor}
\textbf{0.1}     & \textbf{0.1}     & \textbf{62.3} \\
0.1     & 0.5     &  62.0             \\ 
0.5     & 0.1     &  61.7              \\
0.5     & 0.5     & 61.8  \\ \hline
\end{tabular}
\end{table}

\section{Mask Generation}
\label{subsec: mask}
In this section, we delve into the detailed process of creating masks from seeds (see Section.~\ref{subsec: maskgenerate}), and provide an in-depth explanation of how seeds are fused to generate masks (Section.~\ref{subsec: seedfusion}).

\subsection{Modifing Mask Generation}
\label{subsec: maskgenerate}

Thanks to IRNet~\cite{irnet}, we were able to enhance the training of our model's seeds by subsequently refining them. However, our approach did not utilize the IRNet model~\cite{irnet} in its original form; we introduced certain modifications to tailor it more fittingly to our seed. IRNet~\cite{irnet} conventionally sets a threshold to differentiate between the background and foreground before training the model, with this threshold being determined as a hyperparameter. If a value exceeds the foreground threshold, it is classified as foreground, whereas values lower than the background threshold are considered background. Values that fall in between these thresholds are categorized as unknown, and it is in this context that IRNet is trained. In our approach, we chose to dynamically compute this threshold, using values that we derived based on our model.

Inspired by TS-CAM~\cite{TS-CAM}, it is stated that in the ViT, the class token of the attention matrix aids in extracting both the foreground and background (referred to in the paper~\cite{TS-CAM} as the \textit{semantic-agnostic attention map}). We utilized this approach to set the background and foreground thresholds for training IRNet~\cite{irnet}. The background threshold was determined by assigning a certain weight to the \textit{semantic-agnostic attention map} segment, and we established the foreground and background thresholds with an approximate gap of $0.5$ between them. This adjusted thresholding strategy contributed to an improvement in our results by an approximate margin of \textbf{+0.4} over previous outcomes.
\newpage
\clearpage

\subsection{Seed Fusion for Mask Generation}
\label{subsec: seedfusion}
\begin{figure}[t!]
    \centering
    \includegraphics[width=0.7\columnwidth]{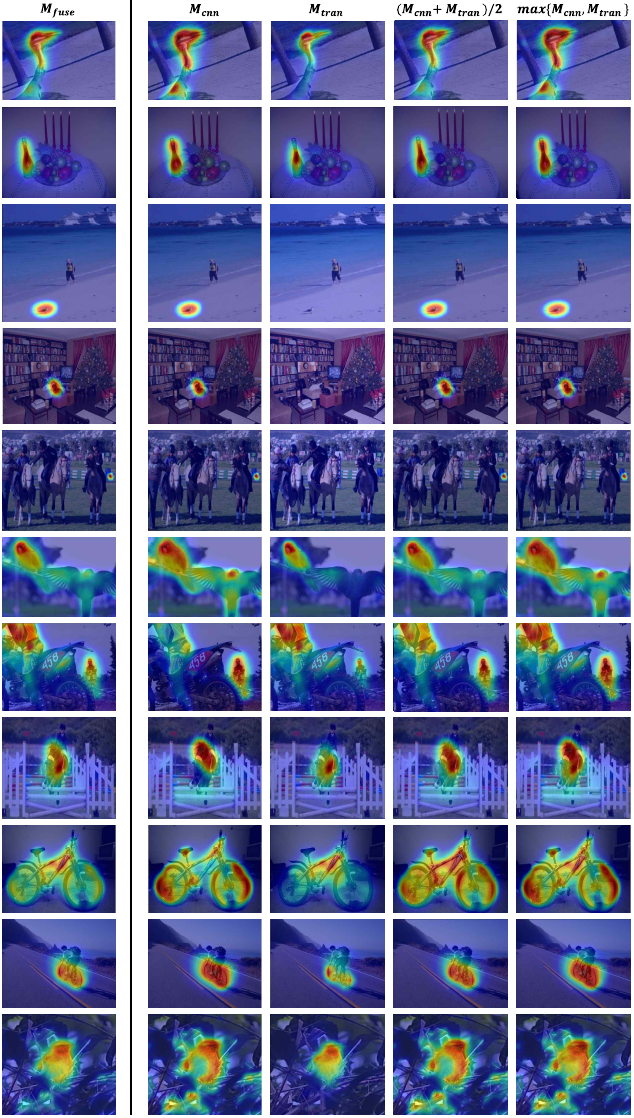}
    \caption{The activation map of CoBra source combination. The best label to be used in the training process for generating a mask is $\mathbf{M}_{fuse}$. $\mathbf{M}_{cnn}$ and $\mathbf{M}_{tran}$ are derived from the CAK branch and SAK branch, respectively. Our $\mathbf{M}_{fuse}$ is $max$ values of $(\mathbf{M}_{cnn} + \mathbf{M}_{tran})/2$ and $\mathbf{M}_{tran}$. Appropriate fusion enables the $\mathbf{M}_{cnn}$ and $\mathbf{M}_{tran}$ to complement each other.}
    \label{fig:sup_fig2}
\end{figure}

\begin{figure}[t!]
    \centering
    \includegraphics[width=0.80\columnwidth]{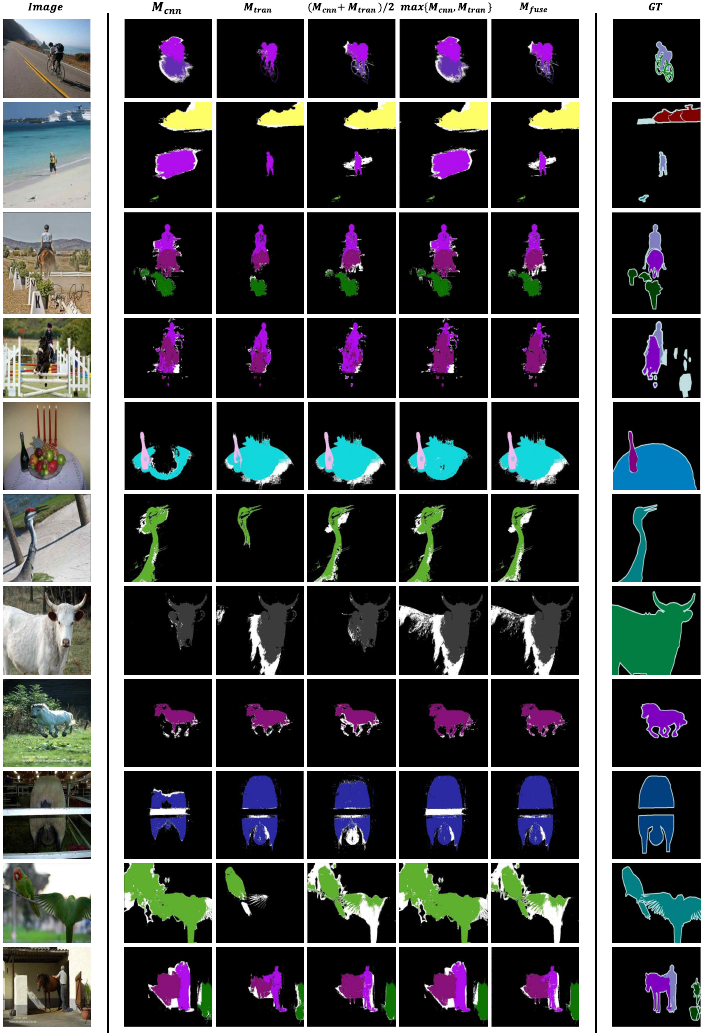}
    \caption{The ir label used in the training process to create a mask. From left to right, the components are the image, $\mathbf{M}_{cnn}$, $\mathbf{M}_{tran}$, $average$ of $\mathbf{M}_{cnn}$ and  $\mathbf{M}_{tran}$, $max$ values of $\mathbf{M}_{cnn}$ and $\mathbf{M}_{tran}$,  $\mathbf{M}_{fuse}$ that is used for our mask, and GT. The black color on the label represents the background, the colored part represents the class, and the white part is an unknown that is difficult to refer to as the background or foreground.}
    \label{fig:sup_fig3}
\end{figure}

The fusion seed and mask are calculated as $\mathbf{M}_{fuse} = \max \left\{ (\mathbf{M}_{cnn}+\mathbf{M}_{tran})/{2}, \hspace{2pt} \mathbf{M}_{tran}\right\}$. Seed Fusion and Mask Fusion are illustrated in Fig.~\ref{fig:sup_fig2} and Fig.~\ref{fig:sup_fig3}. As can be seen in the main paper (Table.5), the value of $\mathbf{M}_{fuse}$ is higher than any other combination. As seen in Fig.~\ref{fig:sup_fig2} demonstrates the fusion method for creating a better seed using CAM obtained from each branch. The second and third column of the Fig.~\ref{fig:sup_fig2} shows $\mathbf{M}_{cnn}$ and $\mathbf{M}_{tran}$ compensate each localization map.
In particular, as seen in the sixth row of Fig.~\ref{fig:sup_fig2}, while $\mathbf{M}_{cnn}$ segments both \texttt{birds}, the $\mathbf{M}_{tran}$ only segments one \texttt{bird}. Notably, $\mathbf{M}_{tran}$ can detect more precisely than $\mathbf{M}_{cnn}$. Just to mention, in case it's not clear, the third, fourth, and fifth rows represent the localization map for \texttt{bird}, \texttt{dog}, and \texttt{chair}, respectively. Since $\mathbf{M}_{tran}$ is trained on patches, its detection capability for small objects is relatively weak. $\mathbf{M}_{cnn}$ compensates for this aspect. By using our $\mathbf{M}_{fuse}$, it is possible to detect missing parts as in the examples mentioned above. 

Fig.~\ref{fig:sup_fig3} shows the label used in the training process to create a mask which is discussed in Section.~\ref{subsec: maskgenerate}. The colored parts, black and white areas represent the foreground, background, and unknown. These labels were created from the CoBra framework using the fusion method described earlier. Our fusion method aims to detect more accurately than others. In practice, owing to our approach for compensating missed elements, we successfully identified areas that could have been missed. Moreover, when excessive regions were detected, adjustments were made to accurately outline the boundaries.

\section{Experimental Setup}
\label{sec: 1}
As we added a loss function to properly integrate CoBra for training, we naturally adjusted the number of training epochs to 20 with 32 batch size. During the training of CoBra, $\lambda_1$ and $\lambda_2$ in the loss function are set to $0.1$ (see Eq.~\eqref{sup_eq1}). As illustrated in Table.~\ref{table:sup_tab11}, we experimented with various values of lambda to achieve the best score. We resized input images to $224 \times 224$. Localization maps are created by summing the output values generated from multiple scales (0.5, 0.75, 1.0, 1.25, 1.5, 1.75, 2.0). The localization map is interpolated in two different sizes, which is then used during the inference process and when refining the localization map through the IRN \cite{irnet}, called $\mathbf{M}_{fuse}$ (more details on Sec.~\ref{subsec: mask}).

To create a mask (refined seed), the IRN \cite{irnet} is trained with a 32 batch size, 512 sizes of cropped images, and 4 training epochs. After conducting a series of experiments on the embedding sizes used in $\mathcal{L}_{sap}$ and $\mathcal{L}_{cap}$, we discovered that the optimal size was 256 in Table.~\ref{table:table2_embed}, which demonstrated the best performance. The method for setting the threshold was inspired based on \cite{threshold}, which provides an insightful analysis of thresholding issues.

\section{Discussion on the Methodology for Determining Thresholds}
\label{sec: thres}

In the task of weakly supervised semantic segmentation, setting a threshold to distinguish the foreground is a common practice~\cite{mctformer, irnet,spatialBCE, threshold}. Typically, this involves specifying a value within the 0 to 1 range of Class Activation Maps (CAMs) to define the background. However, this approach often encounters several issues. A notable problem arises from the varying confidence levels exhibited by different classes in the CAMs. For instance, the class \texttt{chair} might be confused with a \texttt{desk}, and its slender parts like legs result in generally lower confidence scores. By setting a lower background threshold, CAMs can more effectively identify the \texttt{chair} object. 
This means that the threshold can differ for each class. Another challenge is the need for a universal threshold for all images. Just as the foreground values vary across images, the background values can also differ; however, current thresholding methods do not adequately account for this variability. Some of the papers~\cite{amn, threshold} have discussed these issues with thresholding methods. We too carefully contribute to this discussion, proposing a novel approach to threshold determination. Drawing inspiration from TS-CAM~\cite{TS-CAM}, we considered the class token of the attention matrix indicated as the boundary of background and foreground, which can be found in the SAK branch in our model. We could derive the background threshold from this. The degree of weighting given to this factor was determined similarly to conventional threshold-setting methods. This approach has been shown to yield better results compared to traditional methods, demonstrating the effectiveness of our tailored thresholding strategy in weakly supervised semantic segmentation.

\newpage
\clearpage

\end{document}